\documentclass[11pt]{article}

\usepackage[final]{acl}

\usepackage{times}
\usepackage{latexsym}

\usepackage[T1]{fontenc}
\usepackage[utf8]{inputenc}

\usepackage{microtype}

\usepackage{inconsolata}

\usepackage{graphicx}
\usepackage{amsmath}
\usepackage{booktabs}
\usepackage{multirow}
\usepackage[table]{xcolor}
\usepackage{tikz}
\usepackage{pgfplots}
\pgfplotsset{compat=1.18} % or 1.17 if Overleaf complains

\usepackage{booktabs}
\usetikzlibrary{patterns}
\usetikzlibrary{arrows.meta}
\usepackage{booktabs}
\usepackage{array}
\usepackage{caption}
\usepackage{siunitx}
\usepackage{rotating} % for better header handling if needed (optional)
\usepackage{amsfonts}
\usepackage{enumitem}
\usepackage[table]{xcolor}

\definecolor{goldctx}{RGB}{245,248,252}   % very light blue
\definecolor{selfctx}{RGB}{252,248,242}   % very light orange
\definecolor{qonly}{RGB}{248,248,248}     % very light gray

\newcommand{\data}{\texttt{ChronoScope}}
\title{Evaluating Temporal Consistency in Multi-Turn Language Models}

\author{Yash Kumar Atri \quad Steven L. Johnson \quad Tom Hartvigsen \\
  University of Virginia \\
  \texttt{\{atri, steven, hartvigsen\}@virginia.edu}}
  
\begin{document}
\maketitle

\begin{abstract}
Language models are increasingly deployed in interactive settings where users reason about facts over time rather than in isolation.
In such scenarios, correct behavior requires models to maintain and update implicit temporal assumptions established earlier in a conversation.
We study this challenge through the lens of \emph{temporal scope stability}: the ability to preserve, override, or transfer time-scoped factual context across dialogue turns.
We introduce \data, a large-scale diagnostic benchmark designed to isolate temporal scope behavior in controlled multi-turn interactions, comprising over one million deterministically generated question chains grounded in Wikidata.
\data\ evaluates whether models can correctly retain inferred temporal scope when follow-up questions omit explicit time references, spanning implicit carryover, explicit scope switching, cross-entity transfer, and longer temporal trajectories.
Through extensive evaluation of state-of-the-art language models, we find that temporal scope stability is frequently violated in controlled multi-turn settings, with models often drifting toward present-day assumptions despite correct underlying knowledge.
These failures intensify with interaction length and persist even under oracle context conditions, revealing a gap between single-turn factual accuracy and coherent temporal reasoning under sequential interaction.
We make our dataset and evaluation suite publicly available at \url{https://github.com/yashkumaratri/ChronoScope}.
\end{abstract}

 \begin{figure}[!t]
    \centering
    \scalebox{0.46}{
    \includegraphics[trim={3.8cm 2.cm 2cm 1cm},clip]{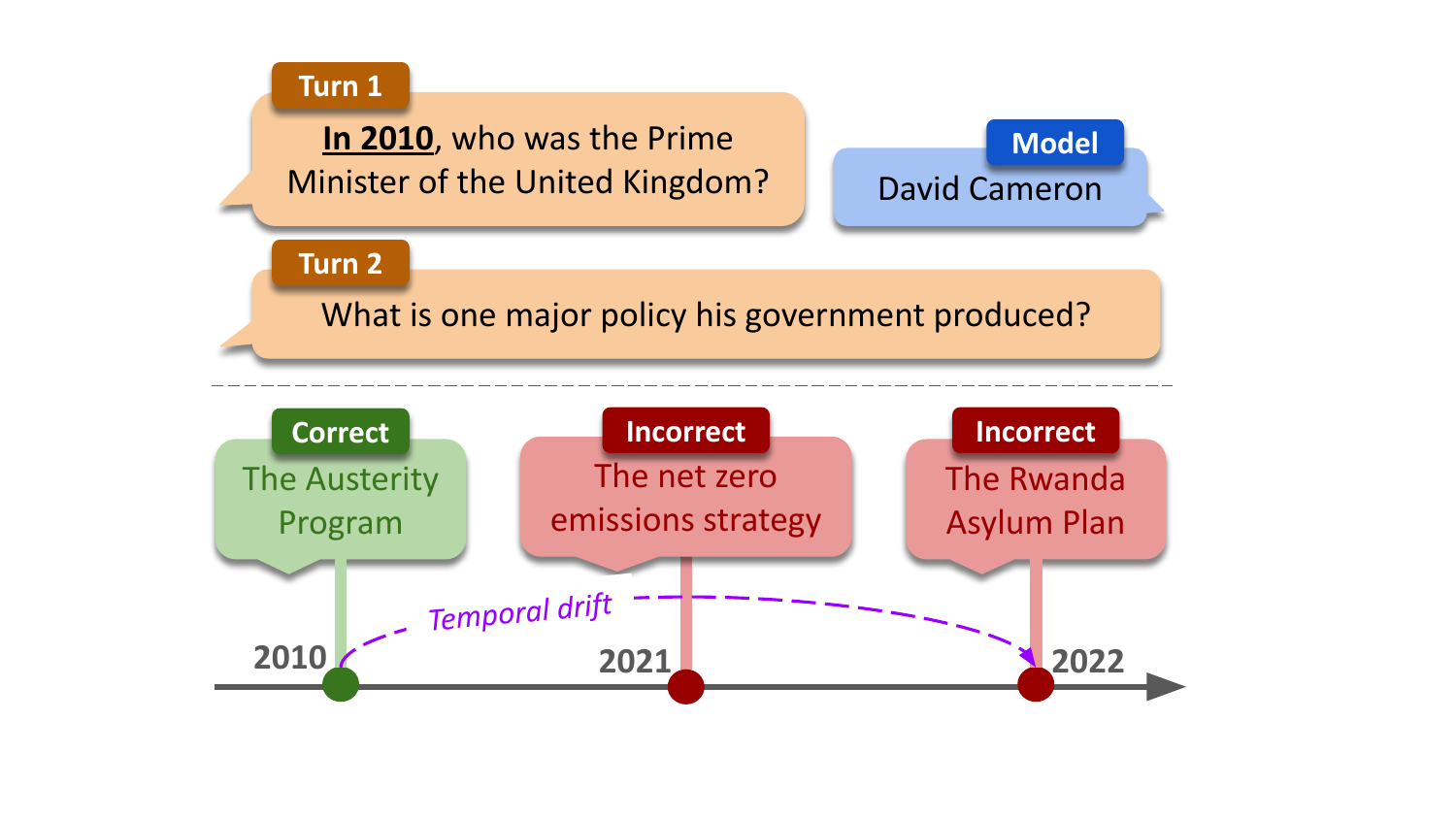}}
    \vspace{-4mm}
    \caption{Illustration of temporal scope drift in multi-turn question answering.
After correctly establishing a historical context (2010), the model answers a follow-up question using factually correct but temporally misaligned policies from later years.}
    \label{fig:modelArch}
    % \vspace{-6mm}
\end{figure}

\section{Introduction}
\label{sec:introduction}

Human conversation is governed by a principle of \emph{contextual persistence}: once a topic, entity, or temporal frame is established, it is assumed to remain valid until a shift is explicitly signaled~\cite{10.1145/3701716.3717744}.
This implicit contract allows speakers to omit redundant constraints while maintaining coherent interaction~\cite{marín2025empiricalcharacterizationtemporalconstraint,ding-etal-2025-tcp}.
As language models are increasingly deployed in sequential, multi-turn settings~\cite{li-etal-2025-firm,chen-etal-2024-efficient}, correct behavior requires them to retain and manipulate contextual assumptions introduced earlier in the conversation~\cite{alazraki-etal-2025-need,dherin2025learningtrainingimplicitdynamics}. This phenomenon has long been studied in discourse and temporal semantics, where conversational interpretation depends on maintaining a shared temporal reference point across utterances. Classical work distinguishes between speech time, event time, and reference time~\cite{ReichenbachHans1947Eosl}, and formal discourse models such as Discourse Representation Theory capture how these reference points evolve across sentences~\cite{kamp1993discourse}.

While prior work has extensively studied factual accuracy and temporal knowledge in single-turn settings \cite{atri-etal-2023-promoting}, far less is understood about how temporal assumptions are maintained, overridden, or transferred across dialogue turns.
Recent studies suggest that even high-performing language models struggle with context inheritance in multi-turn interaction~\cite{laban2025llmslostmultiturnconversation,chenghaozhu-etal-2025-llm}.
A fundamental yet underexplored instance of this challenge is \emph{temporal scope stability}, defined as the ability to preserve, override, or transfer time-scoped factual assumptions across multiple turns~\cite{chenghaozhu-etal-2025-llm,wang2023mint}. In this work, we operationalize temporal scope stability as the ability of a model to maintain an implicit reference time across turns unless an explicit temporal update is introduced.

In natural interaction, users rarely restate temporal constraints explicitly.
Instead, temporal context is typically introduced once and assumed to persist implicitly across follow-up questions~\cite{10.1145/3701716.3717744}.
For example, after asking who served as a country’s prime minister in 2010, a user may naturally inquire about that individual’s policies or actions without repeating the year.
Although such follow-ups are coherent for humans, language models frequently fail to respect the established temporal frame, reverting to default or present-day knowledge~\cite{chenghaozhu-etal-2025-llm,10.1007/978-3-031-82346-6_5}.
These errors do not arise from missing facts, but from a breakdown in contextual coherence, where models answer correctly in isolation yet fail to carry forward an inferred temporal scope~\cite{laban2025llmslostmultiturnconversation}.

Crucially, these failures occur without any parameter updates or external interventions.
The model’s underlying knowledge remains unchanged between turns, yet its interpretation of the query shifts across the interaction~\cite{laban2025llmslostmultiturnconversation}.
This reveals a gap between single-turn factual accuracy and reliable multi-turn reasoning, suggesting that correct knowledge retrieval alone is insufficient for coherent interactive behavior~\cite{sun-etal-2019-dream,reddy-etal-2019-coqa}.
Figure~\ref{fig:modelArch} illustrates a representative failure case.

To address this gap, we introduce \data, a large-scale diagnostic benchmark designed to isolate temporal scope behavior in controlled multi-turn question answering.
\data\ comprises over 1.4 million deterministically generated question chains grounded in Wikidata’s temporally annotated knowledge graph~\cite{10.1145/2629489}.
Each chain begins with a time-scoped factual query and is followed by turns requiring implicit scope carryover, explicit scope switching, or transfer of temporal context across related entities.
By operating over Wikidata entity identifiers, our pipeline enables precise disambiguation and explicit annotation of \emph{temporal drift}, where models produce factually correct but temporally misaligned answers.

Through extensive evaluation of state-of-the-art open-weight and proprietary language models, we find that temporal scope stability is frequently violated in controlled multi-turn settings.
While models perform well when temporal constraints are restated explicitly, accuracy degrades sharply when temporal scope must be inferred and maintained implicitly.
In many failure cases, predicted answers match present-day ground truth, indicating present-day bias rather than general factual confusion.
These failures intensify with interaction length and persist even under oracle conversational context, where prior turns are replaced with gold answers, highlighting temporal scope instability as an inference-time failure mode.

Viewed through the lens of continual learning, temporal scope stability can be seen as an inference-time analogue of continual knowledge retention.
Just as continual learning aims to incorporate new information without overwriting prior knowledge, coherent interaction requires models to temporarily bind knowledge to a contextual frame without reverting to default assumptions.
Evaluating this property is therefore essential for reliable performance in time-sensitive and interactive applications.

Our contributions are as follows:

\begin{enumerate}[leftmargin=*,noitemsep,topsep=0.5pt]
    \item We formalize \emph{temporal scope stability} as a measurable property of multi-turn language model behavior.
    \item We introduce \data, a large-scale diagnostic benchmark comprising over 1.4 million temporally grounded question chains with explicit drift annotations.
    \item We show that state-of-the-art language models often fail to maintain temporal scope across dialogue turns even when the underlying knowledge is correct.
    \item We release the dataset and evaluation code to support future research on temporal and contextual stability in language models.
\end{enumerate}

\begin{table*}[!t]
\centering
\footnotesize
\renewcommand{\arraystretch}{1}
\resizebox{\textwidth}{!}{%
\begin{tabular}{l l c c c l l r}
\toprule
\multirow{2}{*}{\textbf{Dataset}}
& \multirow{2}{*}{\textbf{Source}}
& \multicolumn{3}{c}{\textbf{Temporal Properties}}
& \multicolumn{2}{c}{\textbf{Interaction}}
& \multirow{2}{*}{\textbf{Scale}} \\
\cmidrule(lr){3-5} \cmidrule(lr){6-7}
& 
& \textbf{Temporal}
& \textbf{Multi-}
& \textbf{Scope}
& \textbf{Temporal}
& \textbf{Failure}
&  \\
& 
& \textbf{Qs}
& \textbf{Turn}
& \textbf{Persist}
& \textbf{Signal}
& \textbf{Focus}
&  \\
\midrule

TempQuestions
& KB
& Yes & No & No
& Explicit
& Time grounding
& 1K \\

TimeQA
& Wiki
& Yes & No & No
& Explicit
& Fact staleness
& 17K \\

TimeR1
& Wiki
& Yes & No & No
& Explicit
& Past--present shift
& 200K \\

PAT-Questions
& Wiki
& Yes & No & No
& Explicit
& Present bias
& 6K \\

TEMPTABQA
& Table
& Yes & No & No
& Explicit
& Table alignment
& 11K \\

UnSeenTimeQA
& Synthetic
& Yes & No & No
& Explicit
& Temporal generalization
& 1.6K \\

MultiHop-QA
& Wiki
& Partial & Yes & No
& Incidental
& Depth errors
& $>$10K \\

\midrule
\rowcolor{gray!10}
\textbf{\data\ (Ours)}
& \textbf{KG}
& \textbf{Yes}
& \textbf{Yes}
& \textbf{Yes}
& \textbf{Implicit}
& \textbf{Temporal drift}
& \textbf{1.46M} \\
\bottomrule
\end{tabular}%
}

\caption{
Comparison of temporal question answering benchmarks by temporal supervision and interaction structure.
Most prior datasets rely on Wikipedia-based sources and explicit temporal signals,
whereas \data\ is derived from the Wikidata knowledge graph (KG) and
explicitly evaluates implicit temporal scope persistence and drift under multi-turn interaction.}

% \vspace{-4mm}
\label{tab:chronoscope_comparison}
\end{table*}

\section{Related Work}

Prior work on temporal question answering~\cite{lazaridou2021mindgapassessingtemporal} has largely focused on single-turn queries in which temporal constraints are explicitly specified.
Benchmarks such as TempQuestions~\cite{10.1145/3184558.3191536}, TimeQA~\cite{chen2021datasetansweringtimesensitivequestions}, and TimeR1 \cite{liu2025timer1comprehensivetemporalreasoning} evaluate time-conditioned fact retrieval and reasoning.
However, as noted by~\citet{lazaridou2021mindgapassessingtemporal}, these datasets resolve temporal scope independently for each query and do not assess whether models can maintain a coherent temporal state once explicit cues are removed (Table~\ref{tab:chronoscope_comparison}).

A related line of work examines how language models handle factual change over time and documents a strong present-day bias, where models struggle to override knowledge acquired during pretraining~\cite{Dhingra_2022,10.1145/3477495.3531734,liu2025timer1comprehensivetemporalreasoning}.
In these settings, temporal modifiers are typically provided at every turn, leaving open the question of whether such bias induces contextual drift during multi-turn interaction.
Separately, research on multi-turn question answering has emphasized entity tracking, coreference resolution, and dialogue state consistency~\cite{li-etal-2019-entity,su-etal-2019-improving,wang-etal-2025-rethinking-stateful}, with evidence that strong single-turn performance does not reliably translate to coherent multi-turn behavior~\cite{wang2023mint,laban2025llmslostmultiturnconversation,wang2023mint}.
Interaction-focused benchmarks like Parrot~\cite{sun-etal-2024-parrot} and MTSA~\cite{guo-etal-2025-mtsa} further expose failures in underspecified follow-up queries, while existing multi-turn datasets~\cite{yang-etal-2018-hotpotqa,zhang-etal-2024-end,banatt2024wiltmultiturnmemorizationrobustinductive} assume temporally stable facts.

Beyond recent NLP benchmarks, the importance of maintaining temporal reference points has long been recognized in discourse and linguistic theory. Classical work distinguishes between speech time, event time, and reference time as the core components of temporal interpretation~\cite{ReichenbachHans1947Eosl}. Formal discourse models such as Discourse Representation Theory further describe how these temporal reference points evolve across sentences in a coherent discourse structure~\cite{kamp1993discourse}.

Our work targets an underexplored intersection of these areas by studying \emph{temporal scope stability} under sequential interaction.
We consider multi-turn question chains in which temporal context is introduced by an initial time-scoped query and must be preserved, overridden, or transferred across subsequent turns, often without restating temporal markers~\cite{laban2025llmslostmultiturnconversation,cheng2025temporalblindnessmultiturnllm}.
Unlike prior temporal QA benchmarks, temporal scope is not specified at every turn, and unlike prior conversational QA benchmarks, factual correctness is time-dependent.
This formulation isolates a failure mode not captured by existing datasets: systematic drift toward default or present-day assumptions despite an established temporal context~\cite{liu2025timer1comprehensivetemporalreasoning}.
\data\ therefore provides a diagnostic benchmark for studying temporal coherence under controlled multi-turn interaction.

\begin{table*}[t]
\centering
\footnotesize
\renewcommand{\arraystretch}{1}
\setlength{\tabcolsep}{3pt}
\begin{tabular}{l r r r r r r c}
\toprule
\textbf{Chain Family} 
& \textbf{\#Chains} 
& \textbf{Med Len} 
& \textbf{Max Len} 
& \textbf{Avg Scope Shift} 
& \textbf{Implicit Turns (\%)} 
& \textbf{Span (yrs)} 
& \textbf{Temporal State} \\
\midrule

Carryover              
& 552,408 & 2 & 2  & 0.0 & 51 & 0   & Fixed \\

Carryover-then         
& 324,637 & 2 & 2  & 0.0 & 50 & 0   & Fixed \\

Scope switch           
& 233,098 & 2 & 2  & 1.0 & 6  & 6   & Override \\

Cross-entity then      
& 144,618 & 2 & 2  & 0.0 & 50 & 0   & Fixed \\

Multi-turn chain       
& 72,009  & 3 & 6  & 0.0 & 67 & 2   & Cross-turn \\

Temporal narrative     
& 30,335  & 3 & 11 & 2.3 & 3  & 18  & Evolving \\

Interval reasoning     
& 30,335  & 4 & 5  & 3.0 & 4  & 12  & Aggregative \\

Interval change        
& 27,655  & 5 & 6  & 4.0 & 23 & 15  & Interval \\

Distinct count         
& 27,655  & 4 & 5  & 3.1 & 4  & 10  & Aggregative \\

Change point           
& 24,357  & 4 & 5  & 3.1 & 29 & 8   & Override \\

Bridged multi-pid      
& 2,521   & 4 & 4  & 2.7 & 12 & 5   & Branched \\

\midrule
\textbf{Total}         
& \textbf{1,469,628} &  & 11 &  &  &  &  \\
\bottomrule
\end{tabular}
\caption{
Structural characterization of \data\ chain families.
Implicit Turns denotes the fraction of turns without explicit temporal references.
Span reports the median absolute temporal distance (in years) from the anchor date.
Temporal State summarizes how temporal scope evolves across the chain.
}

% \vspace{-4mm}
\label{tab:chronoscope_final}
\end{table*}

\section{Dataset: \data}
\label{sec:dataset}

\data\ is a large-scale benchmark for evaluating temporal scope stability in multi-turn factual question answering.
It targets a specific failure mode: models abandoning or overwriting an established temporal context across dialogue turns.
Unlike prior temporal QA benchmarks, \data\ treats temporal scope as a latent conversational state inferred from earlier turns and preserved, overridden, or updated as interaction unfolds.

\subsection{Overview}

Each example in \data\ is a short interaction chain consisting of multiple question--answer turns grounded in temporally scoped Wikidata facts.
A chain begins with an explicitly time-scoped question that establishes temporal context, followed by one or more turns that may omit temporal references or introduce updates.
Correct answers therefore depend on both factual knowledge and maintaining the correct temporal scope across turns.

\data\ contains 1{,}469{,}628 chains spanning domains such as politics, business, sports, academia, and public institutions.
Table~\ref{tab:chronoscope_final} summarizes the structural properties of the dataset and its chain families.

\subsection{Data Construction}

\data\ is constructed deterministically from Wikidata snapshots and contains no human-written or model-generated content.
All answers are derived directly from temporally scoped structured knowledge, ensuring correctness and reproducibility.
Data construction proceeds in two stages.
First, we materialize temporally anchored truth tables from Wikidata claims.
Second, we deterministically convert these tables into natural-language questions and multi-turn interaction chains.
Full implementation details are provided in Appendix~\ref{sec:data_construction_details}.

\paragraph{Anchored Truth Construction.}
For each snapshot year, we define a fixed set of anchor dates.
At each anchor, we construct a truth table containing facts valid at that anchor date.
Claims are filtered to remove deprecated statements and self-loops, and temporal qualifiers are parsed from start time, end time, and point-in-time fields.

A claim is considered active at an anchor if its point-in-time matches the anchor year or if the anchor lies within its start--end interval.
For each $(\text{subject}, \text{property})$ pair, we retain a single canonical fact by prioritizing Wikidata ranks and deterministic deduplication.
This yields a stable set of temporally valid facts per anchor date.

\paragraph{Question and Chain Generation.}
Anchored facts are converted into natural-language questions using fixed, property-specific templates parameterized by question type.
Templates generate either explicit temporal questions that mention the anchor year or implicit follow-up questions that omit temporal expressions.

Each interaction chain is anchored to a snapshot year and anchor date, defining the initial temporal scope.
Gold answers correspond to the fact valid at the queried time.
When available, we additionally record the corresponding present-day (2025) answer for the same property to support analysis of present-day bias.

All facts are grouped and deduplicated using Wikidata entity identifiers (QIDs), with English surface labels used only for verbalization.
Multiple interaction patterns are merged into a unified dataset comprising 1{,}469{,}628 interaction chains.

\subsection{Chains and Temporal Scope}

A \emph{chain} is a sequence of question--answer turns $\{(q_1, a_1), \ldots, (q_L, a_L)\}$ grounded in temporally valid Wikidata facts and anchored to a shared temporal context.
Each chain represents a single interaction in which temporal interpretation must be maintained, updated, or transferred across turns.

Temporal scope denotes the effective time frame governing how each turn should be interpreted.
It is introduced explicitly in the first turn via a time-scoped question and may subsequently persist implicitly, be overridden by a new temporal reference, or be transferred across entities.
\data\ evaluates whether models correctly track and apply this evolving temporal scope in multi-turn interactions.

\subsection{Chain Families}

\data\ organizes question chains into 11 \emph{chain families}, each isolating a distinct pattern of temporal scope evolution. Families are programmatically synthesized using Wikidata properties and interaction templates. Table~\ref{tab:chronoscope_final} summarizes their structural properties.

\begin{enumerate}[leftmargin=*,itemsep=1pt,topsep=2pt]
    \item \textbf{Carryover.} A time-scoped first turn followed by an underspecified query that must inherit the scope (e.g., ``In 1998, who was Apple's CEO?'' $\rightarrow$ ``What products did they oversee?'').
    
    \item \textbf{Carryover-Then.} Similar to Carryover but includes discourse connectives like \emph{then} to test if linguistic markers aid or hinder scope inheritance.
    
    \item \textbf{Scope Switch.} The second turn explicitly overrides the prior temporal scope, requiring the model to abandon the previously established year for a new one.
    
    \item \textbf{Cross-Entity Then.} The focus shifts to a related entity while preserving the original temporal scope (e.g., ``In 2012, who was the UK Prime Minister?'' $\rightarrow$ ``Who was the deputy prime minister at that time?'').

    \item \textbf{Multi-Turn Chain.} Chains of 3--6 turns combining implicit carryover with occasional explicit references to test cumulative stability.
    
    \item \textbf{Change Point.} An explicit temporal override occurs after multiple implicit turns, testing the model's ability to detect and update scope transitions.
    
    \item \textbf{Distinct Count.} Requires counting unique values held within a specific window (e.g., ``How many different clubs did Messi play for before 2020?'').
    
    \item \textbf{Interval Reasoning.} Verifies whether a specific fact remains true throughout a continuous temporal span.
    
    \item \textbf{Interval Change.} Combines interval reasoning with explicit updates (e.g., ``Was Obama president in 2008?'' $\rightarrow$ ``What about in 2017?'').
    
    \item \textbf{Temporal Narrative.} Simulates chronological story-tracking with mixed implicit and explicit temporal references.
    
    \item \textbf{Bridged Multi-PID.} Requires composing multiple Wikidata properties under a fixed temporal scope (e.g., ``In 2011, who was the spouse of the French President?'' $\rightarrow$ ``What was their occupation?'').
\end{enumerate}

\begin{table*}[!t]
\centering
\footnotesize
\scalebox{0.805}{
\begin{tabular}{l cccc cccc cccc}
\toprule
\multirow{2}{*}{\textbf{Model}}
& \multicolumn{4}{c}{\textbf{Gold Context}}
& \multicolumn{4}{c}{\textbf{Self-Conditioned}}
& \multicolumn{4}{c}{\textbf{Questions Only}} \\
\cmidrule(lr){2-5} \cmidrule(lr){6-9} \cmidrule(lr){10-13}
& Acc@1 & Final@1 & Chain@1 & Drift
& Acc@1 & Final@1 & Chain@1 & Drift
& Acc@1 & Final@1 & Chain@1 & Drift \\
\midrule
% \rowcolor{gray!10}
Qwen-2.5-7B
& 0.306 & 0.387 & 0.089 & 0.042
& 0.188 & 0.286 & 0.002 & 0.007
& 0.125 & 0.122 & 0.017 & 0.012 \\
Qwen-3-4B
& 0.292 & 0.382 & 0.084 & 0.066
& 0.085 & 0.130 & 0.001 & 0.013
& 0.035 & 0.034 & 0.007 & 0.019 \\
LLaMA-3.1-8B
& 0.253 & 0.306 & 0.059 & 0.022
& 0.171 & 0.249 & 0.002 & 0.008
& 0.028 & 0.012 & 0.008 & 0.017 \\
Mistral-7B
& 0.277 & 0.368 & 0.074 & 0.084
& 0.184 & 0.295 & 0.002 & 0.010
& 0.093 & 0.087 & 0.010 & 0.030 \\
Gemma-7B
& 0.234 & 0.313 & 0.061 & 0.022
& 0.112 & 0.163 & 0.000 & 0.002
& 0.007 & 0.004 & 0.003 & 0.009 \\
DeepSeek-V3
& 0.247 & 0.276 & 0.066 & 0.081
& 0.204 & 0.291 & 0.001 & 0.008
& 0.040 & 0.032 & 0.005 & 0.019 \\
\midrule

ChatGPT-3.5
& 0.323 & 0.384 & 0.032 & 0.226
& 0.188 & 0.226 & 0.008 & 0.284
& 0.114 & 0.092 & 0.006 & 0.346 \\

ChatGPT-4
& 0.441 & 0.516 & 0.061 & 0.163
& 0.294 & 0.353 & 0.015 & 0.215
& 0.192 & 0.164 & 0.012 & 0.292 \\

Gemini-2.5-Flash
& 0.384 & 0.446 & 0.045 & 0.197
& 0.225 & 0.264 & 0.013 & 0.254
& 0.141 & 0.117 & 0.008 & 0.328 \\
\bottomrule
\end{tabular}}
\caption{
\data\ results under three evaluation settings.
Final@1 measures correctness under enforced temporal scope, while Drift captures present-biased substitutions.
All models are evaluated in zero-shot setting. Higher is better for Acc@1, Final@1, and Chain@1, and lower is better for Drift.}

% \vspace{-4mm}
\label{tab:chronoscope_all_settings}
\end{table*}

\paragraph{Annotations, Scale, and Coverage.}
Each chain is annotated with structural attributes (Table~\ref{tab:chronoscope_final}), including family, length, temporal scope updates, and the availability of present-day reference answers.
\data\ comprises 1{,}469{,}628 interaction chains of 2 to 11 turns (median: 2; mean: 2.27).
All questions and answers are generated deterministically from Wikidata snapshots, ensuring exact reproducibility.

\section{Evaluation Methodology}
\label{sec:evaluation}

We evaluate models on \data\ under three complementary context settings designed to isolate distinct aspects of temporal scope reasoning.
Across all settings, the objective is to measure whether a model answers each turn with the temporally correct value while maintaining, overriding, or updating the temporal scope implied by the preceding dialogue.
We report accuracy and drift-to-present rates, and provide full implementation details (prompting, decoding, sampling, and matching) in Appendix~\ref{sec:prompting}.

\subsection{Evaluation Settings}
\label{sec:eval_settings}

Each example is a multi-turn question chain about an entity (or related entities) across time.
The first turn explicitly establishes a temporal scope; subsequent turns may omit time expressions and are evaluated for (i) implicit carryover, (ii) explicit scope switching, (iii) cross-entity transfer, and (iv) longer temporal trajectories.
We evaluate each turn under three settings:

\paragraph{Gold Context.}
The model receives the full conversational history, with all prior assistant responses replaced by the gold answers.
This isolates scope reasoning when the dialogue state is perfectly preserved.

\paragraph{Self-Conditioned.}
The model conditions on its own generated answers from earlier turns.
This reflects realistic interactive usage and captures error propagation and temporal drift induced by self-conditioning.

\paragraph{Questions Only.}
The model answers each turn using only the current question.
This removes conversational state and serves as a lower bound, quantifying reliance on inferred temporal scope rather than explicit temporal cues.

Table~\ref{tab:chronoscope_all_settings} summarizes performance under these settings; detailed analysis is presented in Section~\ref{sec:results}.

\paragraph{Evaluation Metrics. }
We report both turn-level and chain-level metrics.
At the turn level, we measure Acc@1 and surface overlap using token-level and character-level F1.
At the chain level, we evaluate temporal consistency using strict chain accuracy, average per-chain turn accuracy, and final-turn accuracy, which measures whether models recover correct temporal scope by the end of an interaction.

\paragraph{Temporal Drift. }

For a subset of turns, the historically correct answer differs from the present-day answer.
On these \emph{drift-eligible} turns, Drift measures the rate at which models predict the present-day answer instead of the historically correct one, normalized by present-day answer availability.

\paragraph{Aggregation. }

All metrics are reported separately for each evaluation setting and are aggregated over the full dataset, chain families, chain length, and follow-up position.
This enables fine-grained analysis of how temporal scope stability depends on interaction structure.

\paragraph{Models and Baselines.}
We evaluate \data\ on a diverse set of open-weight and proprietary language models,
including Qwen~\cite{qwen2025qwen25technicalreport,yang2025qwen3technicalreport},
LLaMA~\cite{grattafiori2024llama3herdmodels},
Mistral~\cite{jiang2023mistral7b},
Gemma~\cite{gemmateam2025gemma3technicalreport},
and DeepSeek~\cite{deepseekai2025deepseekv3technicalreport},
as well as proprietary models from OpenAI ChatGPT-3.5 and 4~\cite{brown2020languagemodelsfewshotlearners,openai2024gpt4ocard}
and Google Gemini-2.5-Flash~\cite{comanici2025gemini25pushingfrontier}.
All models are evaluated in a zero-shot setting with identical prompts;
implementation and decoding details are provided in Appendix~\ref{sec:baseline_details}.

We report results under three evaluation settings designed to isolate temporal scope reasoning:
\emph{Gold Context}, where oracle historical context is provided;
\emph{Self-Conditioned}, where models must propagate their own answers across turns;
and \emph{Questions Only}, where no prior conversational context is available.
Together, these settings disentangle failures due to temporal scope drift from those due to missing factual knowledge.

\section{Results}
\label{sec:results}

\paragraph{Temporal Scope Enforcement Improves Final-Turn Accuracy.}
\label{subsec:main_results}

Table~\ref{tab:chronoscope_all_settings} summarizes overall turn-level and chain-level performance.
Under Gold Context, all models exhibit a consistent improvement from Acc@1 to Final@1,
with $\Delta = \text{Final@1} - \text{Acc@1}$ ranging from approximately $+0.03$ to $+0.11$.
This pattern indicates that when temporal scope is externally enforced,
models are better able to answer the final query correctly,
even if intermediate turns contain errors.

Because Acc@1 aggregates correctness over all evaluated turns,
it is sensitive to early-turn failures,
whereas Final@1 isolates correctness at the end of each interaction.
The systematic advantage of Final@1 therefore reflects a capability distinct from surface-level factual accuracy.
However, the corresponding Chain@1 scores remain low,
indicating that correct final answers do not imply consistent temporal reasoning throughout the interaction.
This pattern holds across architectures and scales,
suggesting that temporal scope stability is not model-family specific.

\begin{table*}[!t]
\centering
\footnotesize
\renewcommand{\arraystretch}{1.15}
\setlength{\tabcolsep}{4.5pt}
\resizebox{\textwidth}{!}{%
\begin{tabular}{l ccccc ccccc ccccc}
\toprule
\multirow{2}{*}{\textbf{Model}}
& \multicolumn{5}{c}{\textbf{Gold Context (Final@1)}}
& \multicolumn{5}{c}{\textbf{Self-Conditioned (Final@1)}}
& \multicolumn{5}{c}{\textbf{Questions Only (Final@1)}} \\

\cmidrule(lr){2-6}
\cmidrule(lr){7-11}
\cmidrule(lr){12-16}

& L=2 & L=3 & L=4 & L=5 & L=6
& L=2 & L=3 & L=4 & L=5 & L=6
& L=2 & L=3 & L=4 & L=5 & L=6 \\
\midrule

Qwen-2.5-7B
& 0.481 & 0.525 & 0.556 & 0.572 & 0.515
& 0.255 & 0.223 & 0.182 & 0.091 & --
& 0.145 & 0.162 & 0.171 & 0.129 & 0.021 \\

Qwen-3-4B
& 0.482 & 0.530 & 0.561 & 0.585 & 0.540
& 0.117 & 0.101 & 0.084 & 0.034 & --
& 0.037 & 0.041 & 0.043 & 0.028 & 0.006 \\

LLaMA-3.1-8B
& 0.382 & 0.420 & 0.444 & 0.441 & 0.375
& 0.225 & 0.197 & 0.152 & 0.067 & --
& 0.008 & 0.006 & 0.005 & 0.003 & 0.002 \\

Mistral-7B
& 0.458 & 0.501 & 0.532 & 0.548 & 0.452
& 0.263 & 0.231 & 0.190 & 0.081 & --
& 0.108 & 0.121 & 0.128 & 0.109 & 0.014 \\

Gemma-7B
& 0.381 & 0.417 & 0.444 & 0.418 & 0.352
& 0.143 & 0.119 & 0.081 & 0.017 & --
& 0.002 & 0.001 & 0.001 & 0.000 & 0.000 \\

DeepSeek-V3
& 0.319 & 0.342 & 0.361 & 0.332 & 0.251
& 0.257 & 0.229 & 0.186 & 0.064 & --
& 0.039 & 0.044 & 0.047 & 0.048 & 0.009 \\

\midrule

ChatGPT-3.5
& 0.386 & 0.397 & 0.401 & 0.382 & 0.356
& 0.228 & 0.202 & 0.158 & 0.070 & --
& 0.094 & 0.091 & 0.086 & 0.081 & 0.071 \\

ChatGPT-4
& 0.518 & 0.540 & 0.548 & 0.531 & 0.498
& 0.352 & 0.321 & 0.281 & 0.121 & --
& 0.166 & 0.162 & 0.156 & 0.151 & 0.136 \\

Gemini-2.5-Flash
& 0.448 & 0.462 & 0.470 & 0.452 & 0.418
& 0.266 & 0.236 & 0.194 & 0.083 & --
& 0.119 & 0.116 & 0.110 & 0.101 & 0.092 \\

\bottomrule
\end{tabular}%
}

\caption{
Final-turn accuracy (Final@1) as a function of chain length.
Under Gold Context, models benefit from repeated temporal reinforcement up to moderate interaction depths.
In contrast, Self-Conditioned evaluation shows rapid degradation with longer chains due to error propagation,
while Questions Only fails to maintain temporal scope across turns.
Blank entries indicate that no chains of the corresponding length were evaluated in that setting.
}

% \vspace{-5mm}
\label{tab:chronoscope_chain_length}
\end{table*}

\paragraph{Self-Conditioned Evaluation Reveals Rapid State Drift.}
\label{subsec:self_conditioned}

Performance degrades sharply in the \emph{Self-Conditioned} setting, where models must rely on their own prior outputs to maintain context.
Most notably, Strict-Chain Acc@1 collapses to near zero across all tested models, with values ranging from 0.000 to 0.015 (cf. Table~\ref{tab:chronoscope_all_settings}).
This failure indicates that even a single lapse in temporal grounding acts as a catastrophic point of failure,
effectively poisoning the remainder of the interaction and preventing recovery.

Furthermore, Final@1 scores under self-conditioning are consistently lower than those under Gold Context.
While models can exploit externally enforced temporal scope,
they lack robust internal mechanisms for autonomously preserving temporal state.
This failure mode persists across all model families, including large proprietary systems.
Indeed, we observe a \emph{scale--drift paradox}:
models with stronger overall factual performance often exhibit higher drift rates,
suggesting that stronger present-day priors are more difficult to override through implicit conversational context alone.

\paragraph{Removing Context Eliminates Temporal Coherence.}
\label{subsec:questions_only}

The \emph{Questions Only} setting provides a lower bound on temporal scope reasoning.
Without access to prior conversational context, Acc@1 remains low across all models, and Final@1 closely tracks Acc@1.
In this setting, $\Delta$ is near zero or negative,
indicating that final-turn recovery is no longer possible.

These results confirm that \data\ chains cannot be solved by treating each turn independently.
Temporal scope must be inferred from earlier interaction and preserved across turns;
when contextual grounding is removed,
temporal coherence collapses.

\paragraph{Chain Length Exacerbates Temporal Scope Failure.}
\label{subsec:chain_length}

Table~\ref{tab:chronoscope_chain_length} reports Final@1 as a function of chain length.
Under Gold Context, models generally improve with increasing interaction depth up to four or five turns,
reflecting partial accumulation of temporal constraints.
For example, Qwen-2.5-7B improves from $0.48$ at length two to $0.59$ at length five,
while ChatGPT-4 improves from $0.52$ to $0.55$ over the same range.

This trend plateaus or reverses at longer lengths and collapses entirely under Self-Conditioned evaluation.
In self-conditioned chains, Final@1 drops sharply beyond four turns,
often approaching zero,
indicating that temporal scope errors are cumulative rather than isolated.
Additional breakdowns by follow-up position are reported in Appendix~\ref{sec:appendix_results}.

\paragraph{Drift Toward the Present is the Dominant Error Mode.}
\label{subsec:drift}

Across all models and evaluation settings,
present-biased substitution emerges as the dominant error pattern.
Drift(P), which measures the rate at which incorrect answers match valid present-day facts when such facts are available,
ranges from approximately $0.02$ to $0.08$ under Gold Context
and remains substantial under Self-Conditioned evaluation
(Table~\ref{tab:chronoscope_all_settings}).

These errors are not hallucinations.
Instead, models retrieve factually correct answers that are temporally misaligned with the intended scope.
A complete breakdown of performance across all chain families,
including carryover, scope switching, interval reasoning, and temporal narratives,
is provided in Appendix~\ref{sec:appendix_results}.

\begin{table*}[!h]
\centering
\fontsize{8}{10}\selectfont
\renewcommand{\arraystretch}{0.8}
\setlength{\tabcolsep}{1pt}
\begin{tabular}{l c cccc cccc cccc}
\toprule
\multirow{2}{*}{\textbf{Model}} & \multirow{2}{*}{\textbf{Cond.}}
& \multicolumn{4}{c}{\textbf{Gold Context}}
& \multicolumn{4}{c}{\textbf{Self-Conditioned}}
& \multicolumn{4}{c}{\textbf{Questions Only}} \\
\cmidrule(lr){3-6}
\cmidrule(lr){7-10}
\cmidrule(lr){11-14}

& 
& Acc@1 & Final@1 & Chain@1 & Drift
& Acc@1 & Final@1 & Chain@1 & Drift
& Acc@1 & Final@1 & Chain@1 & Drift \\

\midrule

Qwen-2.5-7B & Orig & 0.250 & 0.252 & 0.193 & 0.086 & 0.082 & 0.077 & 0.046 & 0.007 & 0.084 & 0.073 & 0.041 & 0.017 \\
            & Para & 0.250 & 0.254 & 0.195 & 0.113 & 0.082 & 0.079 & 0.048 & 0.008 & 0.083 & 0.073 & 0.041 & 0.014 \\

Qwen-3-4B   & Orig & 0.237 & 0.240 & 0.185 & 0.136 & 0.050 & 0.047 & 0.035 & 0.013 & 0.030 & 0.023 & 0.018 & 0.033 \\
            & Para & 0.235 & 0.240 & 0.186 & 0.201 & 0.055 & 0.049 & 0.035 & 0.010 & 0.037 & 0.032 & 0.023 & 0.034 \\

LLaMA-3.1-8B & Orig & 0.189 & 0.182 & 0.130 & 0.039 & 0.084 & 0.071 & 0.047 & 0.005 & 0.028 & 0.020 & 0.019 & 0.034 \\
             & Para & 0.183 & 0.174 & 0.123 & 0.042 & 0.084 & 0.072 & 0.046 & 0.007 & 0.028 & 0.021 & 0.020 & 0.036 \\

Mistral-7B   & Orig & 0.235 & 0.244 & 0.183 & 0.194 & 0.071 & 0.059 & 0.029 & 0.015 & 0.064 & 0.053 & 0.028 & 0.050 \\
             & Para & 0.237 & 0.248 & 0.188 & 0.207 & 0.071 & 0.060 & 0.031 & 0.015 & 0.061 & 0.050 & 0.025 & 0.044 \\

Gemma-7B     & Orig & 0.204 & 0.216 & 0.159 & 0.026 & 0.037 & 0.024 & 0.014 & 0.004 & 0.007 & 0.007 & 0.007 & 0.019 \\
             & Para & 0.205 & 0.218 & 0.163 & 0.043 & 0.035 & 0.024 & 0.015 & 0.004 & 0.008 & 0.007 & 0.007 & 0.011 \\

DeepSeek-V3  & Orig & 0.217 & 0.219 & 0.169 & 0.182 & 0.071 & 0.056 & 0.028 & 0.023 & 0.030 & 0.022 & 0.015 & 0.033 \\
             & Para & 0.215 & 0.219 & 0.168 & 0.222 & 0.073 & 0.061 & 0.034 & 0.016 & 0.025 & 0.019 & 0.013 & 0.036 \\

ChatGPT-3.5  & Orig & 0.318 & 0.379 & 0.034 & 0.231 & 0.185 & 0.223 & 0.009 & 0.279 & 0.112 & 0.094 & 0.007 & 0.351 \\
             & Para & 0.317 & 0.380 & 0.035 & 0.243 & 0.184 & 0.224 & 0.010 & 0.282 & 0.111 & 0.093 & 0.007 & 0.349 \\

ChatGPT-4    & Orig & 0.438 & 0.512 & 0.059 & 0.168 & 0.291 & 0.349 & 0.016 & 0.219 & 0.189 & 0.161 & 0.013 & 0.296 \\
             & Para & 0.440 & 0.514 & 0.060 & 0.170 & 0.293 & 0.351 & 0.017 & 0.221 & 0.187 & 0.159 & 0.013 & 0.298 \\

Gemini-2.5-Flash & Orig & 0.381 & 0.442 & 0.047 & 0.201 & 0.222 & 0.261 & 0.014 & 0.258 & 0.139 & 0.114 & 0.009 & 0.332 \\
                 & Para & 0.379 & 0.444 & 0.048 & 0.203 & 0.224 & 0.263 & 0.015 & 0.260 & 0.138 & 0.113 & 0.009 & 0.334 \\

\bottomrule
\end{tabular}

\caption{
Performance comparison between the original templated (\textbf{Orig.}) ChronoScope chains and a paraphrased subset (\textbf{Para.}).
Across models, overall performance remains broadly similar across both versions, although drift increases for some models under paraphrasing.
}
\label{tab:chronoscope_paraphrase}
\end{table*}

\paragraph{Robustness to Linguistic Variation.}
\label{subsec:paraphrase}

A potential concern with large-scale synthetic benchmarks is that models may exploit templated surface forms rather than performing genuine reasoning.
To test whether the observed temporal scope failures depend on templated phrasing, we construct a paraphrased subset of \data\ containing 10{,}000 chains.
All user queries are rewritten using the \texttt{gpt-5-2025-08-07} \cite{singh2025openai} model while preserving the underlying entities, temporal anchors, and gold answers.

Table~\ref{tab:chronoscope_paraphrase} compares performance on the original templated chains and the paraphrased subset.
Across models, results remain largely consistent across both versions.
For example, Qwen-2.5-7B achieves nearly identical Gold Context accuracy under both conditions, while drift rates remain comparable and in several cases increase slightly under paraphrasing.
Similar patterns hold across other open-weight and proprietary models.

These results indicate that the temporal drift behavior observed in \data\ is not an artifact of templated surface forms, but reflects a deeper difficulty in maintaining temporal reference points across multi-turn interactions.

\section{Discussion}
\label{sec:discussion}

\data\ reveals a consistent and previously under-measured failure mode in contemporary language models: the inability to reliably maintain temporal scope across multi-turn interactions.

Across all evaluated architectures, models frequently revert to default or present-day assumptions unless temporal constraints are explicitly restated.
Notably, this behavior persists even under Gold Context, where prior turns are replaced with oracle answers (Table~\ref{tab:chronoscope_all_settings}), exposing a gap between surface-level factual competence and coherent temporal reasoning under interaction.

\paragraph{Temporal Scope Stability Is Distinct from Factual Knowledge.}

A primary takeaway of our results is that temporal scope stability is not reducible to factual accuracy.
Across models, final-turn accuracy (Final@1) is consistently higher than turn-level accuracy (Acc@1) under Gold Context (Table~\ref{tab:chronoscope_all_settings}), indicating that models can often recover the correct temporal scope at the end of an interaction even when earlier turns contain errors.
Moreover, Drift rates remain non-trivial even when correct historical answers are available, demonstrating that models often substitute present-day facts in place of temporally appropriate ones.

These errors are not hallucinations.
Instead, models retrieve correct knowledge but bind it to an incorrect temporal frame.
This distinction is critical, as many existing QA benchmarks implicitly conflate knowledge correctness with contextual validity.
\data\ shows that a model may appear reliable in isolation while behaving incoherently under sequential interaction, revealing a blind spot in current evaluation practices.

\paragraph{The Paradox of Model Scale.}

Increased model capacity does not mitigate temporal instability.
In fact, larger proprietary models exhibit some of the highest Drift rates across evaluation settings (Table~\ref{tab:chronoscope_all_settings}).
For example, ChatGPT-4 and Gemini-2.5-Flash achieve strong Acc@1 scores, yet still display substantial Drift under both Gold Context and Self-Conditioned evaluation.

This pattern suggests that stronger memorization of present-day facts during pretraining may amplify resistance to implicit contextual overrides.
Rather than facilitating temporal coherence, scale appears to strengthen default priors that are difficult to suppress through weak conversational signals alone.

\paragraph{Why Self-Conditioning Fails.}

Self-Conditioned evaluation exposes a sharp and consistent collapse in temporal stability across all models.
Chain-level accuracy (Chain@1) approaches zero in nearly every case, even when turn-level accuracy remains moderate (Table~\ref{tab:chronoscope_all_settings}).
This indicates that even minor early deviations rapidly propagate, producing cascading failures from which models rarely recover.

Table~\ref{tab:chronoscope_chain_length} further shows that under Self-Conditioned evaluation, performance collapses beyond very short chains, with Final@1 dropping precipitously at $L \geq 4$.
These trends suggest that temporal scope is not represented as a persistent internal state.
Instead, models appear to infer temporal context opportunistically from local surface cues at each turn.
Once those cues are distorted by self-generated outputs, temporal coherence degrades rapidly.

Importantly, this behavior is consistent across model families, sizes, and training settings, indicating that persistent temporal state tracking is not an emergent property of scale or instruction tuning.

\paragraph{Interaction Length Amplifies Temporal Drift.}

Temporal scope instability compounds with interaction length.
Under Gold Context, increasing chain length initially improves Final@1 by repeatedly reinforcing temporal constraints, but gains saturate and degrade for longer interactions (Table~\ref{tab:chronoscope_chain_length}).
This suggests diminishing returns from repeated supervision when temporal scope is not internally stabilized.

Under Self-Conditioned evaluation, degradation is substantially more severe.
Final@1 collapses beyond short chains, confirming that temporal drift is cumulative rather than a single-turn failure.
These trends mirror challenges observed in long-horizon reasoning and dialogue consistency, indicating that temporal scope instability reflects a broader difficulty in maintaining structured latent state across extended interactions.

\paragraph{Implications for Interactive Systems.}

Temporal drift has direct consequences for real-world deployment.
In domains such as historical analysis, education, policy research, and legal reasoning, users commonly issue iterative queries grounded in a fixed time period.
Present-biased substitutions in these settings can silently corrupt downstream reasoning while remaining superficially plausible.

Our results show that temporal context cannot be assumed to persist unless it is explicitly restated.
Even oracle supervision is insufficient to guarantee temporal coherence across turns.
As a result, reliable interactive systems require stronger guarantees of temporal stability than those provided by current architectures and evaluation protocols.

\paragraph{Toward Temporal Coherence.}

Improving temporal scope stability will likely require interventions beyond standard pretraining and instruction tuning, including explicit temporal state tracking, objectives that reward scope persistence, time-aware retrieval, and decoding strategies that penalize temporal inconsistency across turns. \data\ provides a controlled diagnostic testbed for evaluating such approaches beyond single-turn factual accuracy.

Overall, our findings argue that temporal scope stability should be treated as a first-class evaluation dimension for language models intended for interactive and time-sensitive use.
Without explicit mechanisms to maintain temporal context, improvements in factual accuracy alone are unlikely to yield reliable multi-turn behavior.

\section{Conclusion}
\label{sec:conclusion}

We introduced \emph{temporal scope stability} as a key requirement for language models in interactive, time-sensitive settings and presented \emph{\data}, a large-scale benchmark for evaluating this property in multi-turn factual question answering.
Across models and evaluation regimes, we show that state-of-the-art systems frequently fail to preserve inferred temporal context, drifting toward present-day assumptions even when a clear historical scope is established, without any parameter updates, and that these failures persist under paraphrased variants of the benchmark.
By modeling temporal scope as a latent conversational constraint and providing drift annotations at scale, \data\ exposes a gap between single-turn factual accuracy and coherent multi-turn temporal reasoning and motivates temporal coherence as a first-class evaluation dimension for interactive language systems.

\section{Limitations}
\label{subsec:limitations}

While \data\ provides a rigorous diagnostic for temporal consistency, it has several limitations. First, the benchmark focuses on entity-centric factual questions grounded in Wikidata. It does not currently cover narrative, procedural, counterfactual, or speculative forms of temporal reasoning; thus, it is not intended to evaluate temporal coherence in free-form storytelling or hypothetical scenarios. Second, because the dataset is deterministically generated via templates to ensure factual ground truth across 1.46M chains, it lacks the linguistic variety and pragmatic ambiguity found in natural human-to-human dialogue.

Regarding data freshness, present-day answers are defined relative to a fixed Wikidata snapshot. Although future real-world updates may change which facts qualify as present-day, the core phenomenon under study, temporal drift toward a model’s default or most recent knowledge, remains invariant to the choice of snapshot.

Our evaluation spans a representative set of state-of-the-art language models. We also conducted preliminary experiments with other open-weight systems, including LLaMA-3 variants \cite{grattafiori2024llama3herdmodels}, Phi-4 \cite{abdin2024phi4technicalreport}, and NVIDIA Nemotron \cite{nvidia2024nemotron4340btechnicalreport}. However, these models frequently struggled with the initial zero-shot temporal constraints (Turn 1), leading to high error rates that obscured the analysis of multi-turn \textit{stability}. To maintain a clear focus on how models maintain scope rather than their baseline factual density, we omit these lower-performing baselines from the primary analysis. Their failures, however, reinforce our conclusion that temporal scope stability is not a solved problem even in highly capable model families.

Finally, \data\ identifies the failure mode but does not prescribe a single architectural solution. Addressing temporal scope instability will likely require interventions such as explicit temporal state tracking, multi-turn supervision that penalizes scope violations, or constraint-aware decoding. \data\ serves as the necessary diagnostic testbed to measure progress toward these more reliable, time-sensitive conversational agents.

\section{Ethical Considerations}

\data\ is constructed entirely from Wikidata, a publicly available knowledge base released under a permissive license.
The benchmark contains no private or sensitive personal information beyond widely documented public figures and historical entities.
No human subjects were involved in data collection, annotation, or evaluation.

Our work focuses on diagnosing temporal reasoning failures in language models rather than deploying or modifying such systems.
As such, we do not foresee direct negative societal impacts arising from the release of the benchmark or evaluation code.
However, our findings highlight risks associated with temporal misalignment in deployed language models, particularly in time-sensitive domains such as education, historical analysis, policy research, and legal reasoning, where present-biased errors may lead to misleading or incorrect conclusions.

\section{Acknowledgment}
This work was supported by the McIntire School of Commerce Foundation and by a Faculty AI Research Award from the University of Virginia’s Darden School of Business.

% We release \data\ and associated evaluation code to support transparency and reproducibility, and we encourage future work to consider temporal coherence as a core reliability criterion when developing and deploying interactive language models.

\bibliography{custom}

% \newpage
% \newpage

\clearpage

\newpage

\appendix
\section{Appendix}
This appendix provides full implementation details for all models evaluated on \data\, including decoding settings, prompting protocols, and evaluation metrics.

\subsection{Model and Baseline Details}
\label{sec:baseline_details}

\paragraph{Open-Weight Models.}
We evaluate a diverse set of instruction-tuned open-weight language models spanning multiple architectures and training settings.
These include Qwen-2.5-7B and Qwen-3-4B~\cite{qwen2025qwen25technicalreport,yang2025qwen3technicalreport},
LLaMA-3.1-8B~\cite{grattafiori2024llama3herdmodels},
Mistral-7B-Instruct~\cite{jiang2023mistral7b},
Gemma-7B~\cite{gemmateam2025gemma3technicalreport},
and DeepSeek-V3~\cite{deepseekai2025deepseekv3technicalreport}.
All open-weight models are evaluated using their publicly released instruction-tuned checkpoints without any additional fine-tuning or adaptation.

Inference is performed using greedy decoding with temperature $0.0$, top-$p$ disabled, and a maximum generation length sufficient to cover all answers.
All models are evaluated in a zero-shot setting using identical prompts across evaluation settings.

\paragraph{Proprietary Models.}
To contextualize open-weight performance, we additionally evaluate proprietary models from OpenAI and Google.
These include ChatGPT-3.5 and ChatGPT-4~\cite{brown2020languagemodelsfewshotlearners,openai2024gpt4ocard},
as well as Gemini-2.5-Flash~\cite{comanici2025gemini25pushingfrontier}.
All API-based models are queried using zero-shot prompting without external retrieval, tool use, or system-level memory.

To ensure comparability, proprietary models are evaluated using the same prompt templates as open-weight models.
Reported results for API-based systems should be interpreted as conservative estimates, as decoding parameters and internal state management are not fully controllable.

\paragraph{Additional Evaluated Models.}
We additionally evaluated several widely used instruction-tuned models, including earlier LLaMA-3 variants \cite{grattafiori2024llama3herdmodels}, Phi-3 \cite{abdin2024phi3technicalreporthighly}, Phi-4 \cite{abdin2024phi4technicalreport}, and NVIDIA Nemotron\cite{nvidia2024nemotron4340btechnicalreport}.
Across evaluation settings, these models consistently underperformed relative to the models reported in the main paper, often exhibiting near-zero chain-level accuracy and severe temporal drift even under Gold Context.
For clarity of presentation and space considerations, we exclude these results from the main tables.
Their failure patterns were qualitatively consistent with those discussed in Section~\ref{sec:results}.

\paragraph{Reproducibility.}
All evaluations are deterministic given model outputs and Wikidata-derived gold answers.
We release prompt templates, evaluation scripts, dataset splits, and metadata to support exact reproduction of our results and future benchmarking using \data.

\subsection{Evaluation Details}
\label{sec:eval_details}

This section provides additional details on the evaluation protocol used throughout the paper, complementing the high-level description in Section~\ref{sec:evaluation}.

\subsubsection{Prompting and Context Construction}
\label{sec:prompting}

All models are evaluated using a shared, model-agnostic prompting protocol designed to isolate temporal scope reasoning rather than prompt-following behavior.
Each prompt consists of a system instruction followed by a sequence of user and assistant turns, depending on the evaluation setting.

The system instruction specifies that models should use the provided conversational context, preserve previously established temporal scope unless explicitly changed, and respond with only the requested entity or value.
No additional temporal hints, demonstrations, or reasoning cues are provided.

For a given turn $t$ in a chain, the prompt includes all prior turns $\{1, \ldots, t-1\}$ constructed according to the evaluation setting.
In the \emph{Gold Context} setting, prior answers are replaced with gold answers.
In the \emph{Self-Conditioned} setting, prior turns include the model’s own generated responses.
In the \emph{Questions Only} setting, only the current question is provided.

Prompts are rendered using each model’s native chat template when available; otherwise, a plain-text fallback format with explicit \texttt{User} and \texttt{Assistant} role markers is used.
This ensures consistent role alternation and avoids model-specific prompt engineering.

\paragraph{System Prompt.} All models were provided with the following system instruction to ensure consistency in temporal scope maintenance and output format:

\begin{quote}
\small
\texttt{You are a helpful assistant that answers factual questions.\\
Important:\\
- Use the conversation context.\\
- If a prior turn establishes a time period, keep that same time scope unless it is explicitly changed.\\
- Answer with only the entity or value, no extra explanation.}
\end{quote}

\section{Paraphrasing Procedure}
\label{sec:appendix_paraphrase}

To evaluate robustness to surface-form variation, we construct a paraphrased subset of ChronoScope chains.
Each user query in the selected subset is rewritten using the \texttt{gpt-5-2025-08-07} model while preserving the underlying entities, temporal anchors, and gold answers.

We prompt the model to produce a semantically equivalent question with different surface realization while maintaining the same temporal constraints.
Chains for which the paraphrasing process failed to preserve the original meaning were automatically filtered using answer equivalence checks.

The final paraphrased evaluation set contains 10{,}000 chains spanning all ChronoScope chain families.

\paragraph{Paraphrasing Prompt}
\label{sec:appendix_prompt}

The following prompt template was used to generate paraphrased queries:

\begin{quote}
\small
\texttt{Rewrite the following question so that it has a different surface form while preserving the same meaning, entities, and time references.\\
Original question: [QUESTION]\\
Constraints:\\
- Preserve all entities.\\
- Preserve temporal references and years.\\
- Do not change the factual answer.}
\end{quote}

\paragraph{Evaluation Units.}
Each example in \data\ is a multi-turn question chain $\{(q_t, a_t)\}_{t=1}^{T}$ grounded in Wikidata.
Evaluation is performed at the turn level for all turns $t \geq 2$, unless otherwise specified, as the first turn explicitly establishes temporal scope.
We additionally report chain-level metrics that aggregate turn-level performance across an entire interaction.

\paragraph{Sampling Strategy.}
To ensure balanced coverage across interaction patterns, chains are stratified by (i) chain family and (ii) chain length.
For each experiment, we sample up to $100,000$ chains from each stratum and report the total number of evaluated chains and turns alongside all metrics.
Self-conditioned evaluation is performed on a subset of the sampled chains due to its sequential nature.

\paragraph{Decoding Configuration.}
All open-weight models are evaluated using deterministic decoding with temperature $0.0$ and no nucleus sampling.
We cap generation length to a small fixed budget sufficient to cover all entity and numeric answers.
For top-$k$ evaluation, we generate multiple candidate answers per turn using multiple return sequences and/or beam search, and mark a turn as correct if any candidate matches the gold answer.

\paragraph{Answer Matching.}
A prediction is marked correct if it matches the gold answer under a relaxed string matching procedure.
This procedure normalizes whitespace and punctuation, removes common answer prefixes (e.g., “Answer:”), and allows exact or substring matches after normalization.
We additionally report token-level and character-level overlap scores (micro-averaged) as auxiliary similarity measures.

% When enabled, we also support semantic matching using sentence embeddings with a fixed cosine similarity threshold.

\paragraph{Present-Day Drift Measurement.}
For turns where a present-day reference answer is available, we identify \emph{present-day drift} when a prediction is incorrect with respect to the temporally grounded gold answer but matches the present-day answer under the same matching function.
We report drift rates conditioned on present-day availability, as well as drift rates conditioned on model errors, to isolate systematic bias toward present-day assumptions.

\paragraph{Chain-Level Metrics.}
In addition to per-turn accuracy, we report strict chain accuracy (the fraction of chains with all evaluated turns answered correctly), average per-chain turn accuracy, and final-turn accuracy.
These metrics capture how errors accumulate over interaction length and whether models recover or deteriorate over time.

\paragraph{Determinism and Reproducibility.}
All evaluations are deterministic given model outputs and Wikidata-derived gold answers.
Randomness is limited to chain sampling and is controlled via fixed seeds.
We release all evaluation scripts, prompt templates, dataset splits, and metadata to enable exact reproduction of our results and future benchmarking on \data.

\subsection{Dataset Construction Details}
\label{sec:data_construction_details}

This section describes the construction of \data, including snapshot selection, anchored truth table generation, and deterministic conversion into single-turn and multi-turn question chains.

\subsubsection{Stage 1: Wikidata Snapshot Acquisition}
\label{sec:stage1_snapshots}

We construct \data\ from official Wikidata snapshots corresponding to four reference years: 2022, 2023, 2024, and 2025.
Each snapshot represents a temporally consistent view of the knowledge graph at that time and serves as the basis for all subsequent processing.
Using multiple yearly snapshots allows us to evaluate temporal scope stability across historical contexts while keeping data construction fully deterministic.

\subsubsection{Stage 2: Anchored Truth Table Construction}
\label{sec:stage2_truth}

For each snapshot year, we define a fixed set of temporal anchors (quarterly by default).
At each anchor, we materialize truth tables containing temporal facts that are active at the anchor date.
Facts are filtered to remove deprecated statements and self-loops, and temporal validity is determined using start time, end time, and point-in-time qualifiers.

We begin from Wikidata claims filtered to remove deprecated statements and, optionally, self-loops where the subject and value coincide.
Time qualifiers are parsed from JSON-encoded lists and canonicalized deterministically by selecting the earliest start time, latest end time, and earliest point-in-time when available.
A claim is marked temporal if any time qualifier is present.

For a given anchor, a temporal claim is considered active if either (i) its point-in-time qualifier matches the anchor year, or (ii) the anchor lies within its start--end interval, allowing open-ended intervals.
Timeless claims are processed separately.
For each anchor and for each subject--property pair, we retain a single canonical claim by prioritizing Wikidata ranks (preferred $>$ normal) and deduplicating by $(\text{subject}, \text{property})$.
The resulting truth tables are stored as Parquet files and include selected qualifier fields used in later stages.

\subsubsection{Stage 3: Question and Chain Generation}
\label{sec:stage3_generation}

We deterministically convert anchored truth tables into natural-language questions using fixed, property-specific templates.
Templates are parameterized by question type (e.g., \emph{who}, \emph{what}, \emph{where}, \emph{which}) and generate either explicit temporal questions (including the anchor year or date) or implicit follow-up questions that omit temporal expressions.

Using these templates, we construct both single-turn questions and multi-turn chains.
Chain families include direct factual queries, implicit carryover chains where temporal scope must be preserved across turns, explicit scope-switch chains that override the established temporal context, cross-entity transfer chains that require propagating scope to a related entity, longer trajectories querying multiple properties under a shared scope, and multi-hop and analytic variants that require reasoning over intermediate entities or aggregations.

To ensure linguistic quality, we attach English labels to entities and values and apply conservative label hygiene filters, removing identifiers, malformed strings, and duplicates.
For each chain, we store the temporally grounded gold answer for every turn.
When available, we additionally store a present-day reference answer for the same property, enabling explicit identification of present-day drift during evaluation.

Given fixed truth tables, templates, and random seeds, this generation process is fully deterministic.

\subsection{Additional Discussion}

\label{sec:faq}

\paragraph{Why not evaluate only the final turn?}
Final-turn accuracy captures whether a model eventually recovers the correct temporal scope, but it obscures \emph{when} and \emph{how} temporal drift arises. Our analysis shows that models frequently answer early follow-up turns incorrectly even when they recover later under oracle context, and that these early errors compound under self-conditioned evaluation. Reporting both turn-level and final-turn metrics therefore provides complementary insight into temporal scope stability.

\paragraph{Why is the first turn excluded from evaluation?}
The first turn in each chain explicitly specifies the temporal scope and can typically be answered using single-turn temporal knowledge. Including it would inflate accuracy without testing temporal scope \emph{persistence}. We therefore begin evaluation from the first follow-up turn, where temporal context must be inferred and maintained implicitly, isolating the phenomenon of temporal scope stability.

\paragraph{Why is \data\ constructed from Wikidata rather than text corpora?}
Wikidata provides structured, time-qualified facts with explicit validity intervals, enabling deterministic generation of temporally grounded question chains at scale. Using a knowledge graph avoids ambiguity from textual timestamps, ensures unambiguous gold answers, and allows precise identification of present-day alternatives for drift analysis. This structure is essential for diagnostic evaluation of temporal scope rather than surface-level temporal reasoning.

\paragraph{Are temporal drift errors simply hallucinations or missing knowledge?}
No. In drift cases, models typically produce factually correct answers that correspond to an incorrect temporal scope, most often the present day. We explicitly record present-day answers from Wikidata and measure how often incorrect outputs match these values. The prevalence of such matches confirms that temporal drift reflects unstable temporal binding rather than hallucination or lack of factual knowledge.

\begin{table}[!h]
\centering
\footnotesize
\scalebox{0.850}{
\begin{tabular}{c c c c}
\toprule
Chain Length &
Acc@1 &
Final-Turn Acc@1 &
Drift@1$\mid$Present \\
\midrule
2 & 0.18 & 0.19 & 0.10 \\
3 & 0.13 & 0.13 & 0.07 \\
4 & 0.35 & 0.52 & 0.03 \\
5 & 0.39 & 0.63 & 0.01 \\
6 & 0.32 & 0.62 & 0.02 \\
7 & 0.16 & 0.37 & 0.09 \\
8+ & 0.09 & 0.21 & 0.11 \\
\bottomrule
\end{tabular}}
\caption{Gold-setting accuracy and drift as a function of chain length (Qwen-2.5-7B).}
\label{tab:app_chronoscope_chain_length}
\end{table}

\subsection{Additional Results}
\label{sec:appendix_results}

This appendix provides additional quantitative analyses that complement the main results in Section~\ref{sec:results}. While the main paper focuses on high-level trends and representative settings, the tables below offer a more granular view of temporal scope behavior across models, interaction lengths, and reasoning families.

\begin{table*}[t]
\centering
\footnotesize
\scalebox{1}{
\begin{tabular}{l l r c c c c}
\toprule
Model & Setting & \#Chains &
Acc@1 &
Strict-Chain Acc@1 &
Final-Turn Acc@1 &
Drift@1$\mid$Present \\
\midrule
Qwen-2.5-7B & Gold & 100k & 0.305 & 0.089 & 0.387 & 0.042 \\
Qwen-2.5-7B & Q-only & 100k & 0.125 & 0.017 & 0.122 & 0.012 \\
Qwen-2.5-7B & Self & 20k & 0.188 & 0.002 & 0.286 & 0.007 \\
\midrule
Qwen-3-4B & Gold & 100k & 0.292 & 0.083 & 0.382 & 0.066 \\
DeepSeek-V3 & Gold & 100k & 0.247 & 0.066 & 0.276 & 0.081 \\
Mistral-7B & Gold & 100k & 0.277 & 0.074 & 0.368 & 0.084 \\
LLaMA-3.1-8B & Gold & 100k & 0.253 & 0.059 & 0.306 & 0.022 \\
\bottomrule
\end{tabular}}
\caption{Overall \data\ performance across models and evaluation settings.}
\label{tab:chronoscope_overall}
\end{table*}

\begin{table}[!t]
\centering
\footnotesize
\scalebox{0.75}{
\begin{tabular}{l c c c}
\toprule
Chain Family &
Acc@1 &
Strict-Chain Acc@1 &
Drift@1$\mid$Present \\
\midrule
Interval Reasoning & 0.675 & 0.120 & 0.005 \\
Interval Change & 0.379 & 0.003 & 0.006 \\
Distinct Count & 0.486 & 0.014 & 0.004 \\
Temporal Narrative & 0.172 & 0.040 & 0.075 \\
Carryover & 0.409 & 0.409 & 0.078 \\
Scope Switch & 0.027 & 0.027 & 0.187 \\
\bottomrule
\end{tabular}}
\caption{Gold-setting performance by temporal reasoning family (Qwen-2.5-7B).}
\label{tab:app_chronoscope_family}
\end{table}

\paragraph{Overall Performance Across Models and Settings.}
Table~\ref{tab:chronoscope_overall} reports aggregate performance for all evaluated models under the three evaluation settings: Gold Context, Self-Conditioned, and Questions Only.
In addition to turn-level accuracy (Acc@1), we include strict chain accuracy and final-turn accuracy to characterize temporal consistency across entire interactions.
Consistent with the main findings, strict chain accuracy remains near zero for most models under self-conditioning, while present-biased drift persists even when oracle context is provided.
These results confirm that temporal scope instability is widespread and not confined to a particular model family.

\paragraph{Effect of Chain Length.}
Table~\ref{tab:app_chronoscope_chain_length} stratifies final-turn accuracy by chain length under each evaluation setting.
Under Gold Context, models generally benefit from longer chains up to four or five turns, suggesting that repeated exposure to the same temporal frame can reinforce correct scope.
However, this trend reverses for longer chains and collapses entirely under Self-Conditioned evaluation, where performance drops sharply beyond short interactions.
This supports the conclusion that temporal scope errors are cumulative and difficult to recover from once introduced.

\paragraph{Performance by Temporal Reasoning Family.}
Table~\ref{tab:app_chronoscope_family} breaks down Gold Context performance by chain family.
Families involving explicit temporal overrides (e.g., scope switch and change point) and narrative progression exhibit substantially higher drift rates and lower strict chain accuracy than simple carryover chains.
This analysis shows that \data\ captures multiple distinct failure modes, and that temporal drift is especially pronounced when scope must be implicitly preserved or updated over several turns.

\begin{table}[!h]
\centering
\footnotesize
\begin{tabular}{c c c}
\toprule
Follow-Up Turn &
Acc@1 &
Drift@1$\mid$Present \\
\midrule
Turn 1 & 0.164 & 0.064 \\
Turn 2 & 0.141 & 0.072 \\
Turn 3 & 0.129 & 0.081 \\
Turn 4 & 0.118 & 0.089 \\
Turn 5+ & 0.104 & 0.097 \\
\bottomrule
\end{tabular}
\caption{Temporal drift increases with follow-up depth (Gold setting).}
\label{tab:chronoscope_followup}
\end{table}

\paragraph{Drift as a Function of Follow-Up Depth.}
Table~\ref{tab:chronoscope_followup} reports accuracy and drift rates as a function of follow-up position within a chain.
Drift rates increase monotonically with turn depth, while accuracy declines, indicating that temporal instability compounds over interaction length.
This pattern holds even under Gold Context, reinforcing the conclusion that temporal scope stability is not a single-turn phenomenon but a long-horizon interaction challenge.

Taken together, these additional results reinforce the central claim of the paper: contemporary language models lack robust mechanisms for maintaining latent temporal scope across multi-turn interactions, leading to systematic present-biased errors even when correct historical knowledge is available.

\subsection{Evaluation Metrics}
\label{sec:appendix_evaluation}

\paragraph{Notation.}
Let $\mathcal{T}$ denote the set of evaluated turns and $\mathcal{C}$ the set of evaluated chains.
For turn $t$, let $y_t$ denote the gold answer and
$\hat{y}_t^{(k)} = \{\hat{y}_{t,1}, \ldots, \hat{y}_{t,k}\}$ the model’s top-$k$ predictions.

\paragraph{Turn-Level Accuracy.}
Top-1 and top-$k$ accuracy are defined as:
\begin{equation}
\mathrm{Acc@1} =
\frac{1}{|\mathcal{T}|}
\sum_{t \in \mathcal{T}}
\mathbb{I}[\hat{y}_{t,1} = y_t],
\end{equation}
\begin{equation}
\mathrm{Acc@k} =
\frac{1}{|\mathcal{T}|}
\sum_{t \in \mathcal{T}}
\mathbb{I}[\exists i \le k : \hat{y}_{t,i} = y_t].
\end{equation}

We additionally report token-level F1 and character $n$-gram F1 ($n=3$) to capture partial surface overlap.

\paragraph{Chain-Level Consistency.}
For a chain $c$ with turns $\{t_1, \ldots, t_{m_c}\}$, strict chain correctness is:
\begin{equation}
\mathrm{Strict}(c) =
\prod_{j=1}^{m_c}
\mathbb{I}[\hat{y}_{t_j,1} = y_{t_j}].
\end{equation}

Strict chain accuracy, average per-chain turn accuracy, and final-turn accuracy are defined as:
\begin{equation}
\mathrm{StrictChain@1} =
\frac{1}{|\mathcal{C}|}
\sum_{c \in \mathcal{C}} \mathrm{Strict}(c),
\end{equation}
\begin{equation}
\mathrm{AvgTurnAcc@1} =
\frac{1}{|\mathcal{C}|}
\sum_{c \in \mathcal{C}}
\frac{1}{m_c}
\sum_{j=1}^{m_c}
\mathbb{I}[\hat{y}_{t_j,1} = y_{t_j}],
\end{equation}
\begin{equation}
\mathrm{Final@1} =
\frac{1}{|\mathcal{C}|}
\sum_{c \in \mathcal{C}}
\mathbb{I}[\hat{y}_{t_{m_c},1} = y_{t_{m_c}}].
\end{equation}

\paragraph{Temporal Drift.}
Let $\mathcal{D} \subset \mathcal{T}$ denote turns where a present-day answer $y_t^{\text{pres}}$ exists and differs from $y_t$.
Drift at top-1 is defined as:
\begin{equation}
\mathrm{Drift@1} =
\frac{1}{|\mathcal{D}|}
\sum_{t \in \mathcal{D}}
\mathbb{I}[\hat{y}_{t,1} = y_t^{\text{pres}} \wedge \hat{y}_{t,1} \neq y_t].
\end{equation}

All metrics are reported separately for each evaluation setting and stratified by chain family, chain length, and follow-up position.

\paragraph{Answer Matching and Drift Eligibility.}
All reported metrics use the normalized exact-match procedure described in the main evaluation section.
Predictions are normalized for whitespace, punctuation, and common answer prefixes before comparison with gold answers.
A turn is marked as drift-eligible only when a valid present-day answer exists and is distinct from the temporally correct answer for that turn.
This distinction is important because it ensures that Drift@1 measures a specific form of temporally misaligned retrieval rather than generic incorrectness.
As a result, high drift values indicate present-biased substitution errors rather than hallucinations or parsing failures.

\paragraph{Example ChronoScope Chains.}
Table~\ref{tab:chronoscope_examples_a} and Table~\ref{tab:chronoscope_examples_b} show representative examples from ChronoScope spanning several chain families.
These examples illustrate the core interaction patterns targeted by the benchmark: \emph{carryover} chains require models to preserve an implicitly established temporal frame, \emph{scope-switch} chains require overriding the prior temporal frame when a new year is introduced, \emph{cross-entity} chains transfer temporal scope to a related entity, and \emph{temporal narrative} chains require tracking a sequence of temporally evolving facts over multiple turns.
Several examples are additionally \emph{drift-eligible}, meaning that a valid present-day answer exists but differs from the historically correct one, enabling explicit measurement of present-biased substitution errors.

\begin{table*}[t]
\centering
\footnotesize
\renewcommand{\arraystretch}{1.08}
\setlength{\tabcolsep}{4pt}
\begin{tabular}{p{2.0cm} p{5.2cm} p{4.2cm} p{3.2cm}}
\toprule
\textbf{Family} & \textbf{Question Chain} & \textbf{Gold Answers} & \textbf{Notes} \\
\midrule

Scope Switch
&
\textbf{T1:} Where did Sang-Hun Jeong work in 2014? \newline
\textbf{T2:} What about in 2005: who employed them?
&
\textbf{T1:} Gangneung-Wonju National University College of Dentistry \newline
\textbf{T2:} Korea Basic Science Institute
&
Explicit temporal override at T2; drift-eligible at T1. \\

Cross-Entity Then
&
\textbf{T1:} In 2022, what type of thing was Jelemek? \newline
\textbf{T2:} And for Neuforweiler, how would you categorize it at the time?
&
\textbf{T1:} cadastral area in the Czech Republic \newline
\textbf{T2:} neighborhood
&
Temporal frame transfers across entities. \\

Carryover-Then
&
\textbf{T1:} In 2022, which country was Ishimsky District in? \newline
\textbf{T2:} Where was it located administratively at the time?
&
\textbf{T1:} Russia \newline
\textbf{T2:} Tyumen Oblast
&
Implicit carryover using discourse marker ``at the time.'' \\

Carryover
&
\textbf{T1:} Which organization did Jaakko Astola work for in 2022? \newline
\textbf{T2:} Which organization was it a member of?
&
\textbf{T1:} Research Council of Finland \newline
\textbf{T2:} Academia Europaea
&
Second turn depends on inherited temporal scope; drift-eligible at T2. \\

Temporal Narrative
&
\textbf{T1:} In 1978, which organization was Yemen Football Association a member of? \newline
\textbf{T2:} What about in 1980: what group was it part of? \newline
\textbf{T3:} What about in 2009? \newline
\textbf{T4:} What about in 2016?
&
\textbf{T1:} Union of Arab Football Associations \newline
\textbf{T2:} Asian Football Confederation \newline
\textbf{T3:} West Asian Football Federation \newline
\textbf{T4:} Gulf Cup Federation
&
Multi-turn temporal trajectory over the same subject. \\

\bottomrule
\end{tabular}
\caption{Representative ChronoScope chains illustrating scope switching, cross-entity transfer, implicit carryover, and temporal narratives.}
\label{tab:chronoscope_examples_a}
\end{table*}

\begin{table*}[t]
\centering
\footnotesize
\renewcommand{\arraystretch}{1.08}
\setlength{\tabcolsep}{4pt}
\begin{tabular}{p{2.0cm} p{5.2cm} p{4.2cm} p{3.2cm}}
\toprule
\textbf{Family} & \textbf{Question Chain} & \textbf{Gold Answers} & \textbf{Notes} \\
\midrule

Scope Switch
&
\textbf{T1:} In 2020, which administrative area was Knatten located in? \newline
\textbf{T2:} What about in 2018: where was it located administratively?
&
\textbf{T1:} Holmestrand \newline
\textbf{T2:} Indre Fosen Municipality
&
Explicit year change reverses the temporal frame; drift-eligible at T1. \\

Carryover-Then
&
\textbf{T1:} Which administrative entity contained Borok in 2022? \newline
\textbf{T2:} Which country was it in then?
&
\textbf{T1:} Vereteysky rural settlement \newline
\textbf{T2:} Ukraine
&
Carryover with discourse connective ``then''; drift-eligible at T1. \\

Cross-Entity Then
&
\textbf{T1:} What political party was Carlo Graaff part of in 2022? \newline
\textbf{T2:} And for Horst Achilles, what party was that at the time?
&
\textbf{T1:} Nazi Party \newline
\textbf{T2:} Nazi Party
&
Cross-entity transfer with preserved temporal context. \\

Temporal Narrative
&
\textbf{T1:} Who employed Elsayed M. Zahran in 2011? \newline
\textbf{T2:} In 2013, where did they work? \newline
\textbf{T3:} What about in 2018?
&
\textbf{T1:} University of Miami \newline
\textbf{T2:} University of Miami \newline
\textbf{T3:} Ball State University
&
Illustrates temporal persistence followed by later change; drift-eligible at T3. \\

Carryover
&
\textbf{T1:} What organization did Christoph Bauer belong to in 2022? \newline
\textbf{T2:} Where did they work?
&
\textbf{T1:} SPD-Stadtverband Freising \newline
\textbf{T2:} University of Geneva
&
Implicit carryover across different relations. \\

\bottomrule
\end{tabular}
\caption{Additional ChronoScope examples showing explicit temporal overrides, carryover with discourse markers, and multi-turn temporal narratives.}
\label{tab:chronoscope_examples_b}
\end{table*}

\end{document}